\documentclass{article}

\usepackage{microtype}
\usepackage{graphicx}
\usepackage{subfigure}
\usepackage{booktabs} 
\usepackage{multirow}
\usepackage{hyperref}
\usepackage{enumitem}

\usepackage{listings}
\usepackage{xcolor} 

\usepackage[accepted]{icml2025} %

\usepackage{amsmath}
\usepackage{amssymb}
\usepackage{mathtools}
\usepackage{amsthm}

\usepackage[capitalize,noabbrev]{cleveref}

\theoremstyle{plain}

\theoremstyle{definition}

\theoremstyle{remark}

\usepackage[textsize=tiny]{todonotes}
\usepackage[framemethod=TikZ]{mdframed}
\mdfsetup{%
    middlelinecolor =   none,
    middlelinewidth =   1pt,
    backgroundcolor =   blue!5,
    roundcorner     =   5pt,
}
\newmdenv[
    middlelinecolor=none,
    middlelinewidth=1pt,
    backgroundcolor=blue!5,
    roundcorner=5pt
]{bluebox}

\newmdenv[
    middlelinecolor=none,
    middlelinewidth=1pt,
    backgroundcolor=gray!20,
    roundcorner=5pt
]{graybox}
\usepackage{xcolor}
\usepackage{tcolorbox}
\newtcbox{\grayboxtext}[1][]{
    on line,
    colframe=gray!20,
    colback=gray!20,
    boxrule=0.5pt,
    arc=4pt,
    boxsep=0pt,
    left=2pt,
    right=2pt,
    top=1pt,
    bottom=1pt,
    #1
}

\icmltitlerunning{Reusing Embeddings: Reproducible Reward Model Research in Large Language Model Alignment without GPUs}

\begin{document}

\twocolumn[
\icmltitle{Reusing Embeddings: Reproducible Reward Model Research \\in Large Language Model Alignment without GPUs}



\icmlsetsymbol{equal}{*}

\begin{icmlauthorlist}
\icmlauthor{Hao Sun}{ucam}
\icmlauthor{Yunyi Shen}{mit}
\icmlauthor{Jean-Fran\c cois Ton}{tt}
\icmlauthor{Mihaela van der Schaar}{ucam}
\end{icmlauthorlist}

\icmlaffiliation{ucam}{University of Cambridge}
\icmlaffiliation{tt}{ByteDance Research}
\icmlaffiliation{mit}{Massachusetts Institute of Technology}

\icmlcorrespondingauthor{Hao Sun}{hs789@cam.ac.uk}

\icmlkeywords{Machine Learning, ICML}

\vskip 0.3in
]



\printAffiliationsAndNotice{}  

\begin{abstract}
Large Language Models (LLMs) have made substantial strides in structured tasks through Reinforcement Learning (RL), demonstrating proficiency in mathematical reasoning and code generation. However, applying RL in broader domains like chat bots and content generation --- through the process known as Reinforcement Learning from Human Feedback (RLHF) --- presents unique challenges. Reward models in RLHF are critical, acting as proxies that evaluate the alignment of LLM outputs with human intent. Despite advancements, the development of reward models is hindered by challenges such as computational heavy training, costly evaluation, and therefore poor reproducibility. \textbf{We advocate for using embedding-based input in reward model research as an accelerated solution to those challenges.} By leveraging embeddings for reward modeling, we can enhance reproducibility, reduce computational demands on hardware, improve training stability, and significantly reduce training and evaluation costs, hence facilitating fair and efficient comparisons in this active research area. We then show a case study of reproducing existing reward model ensemble research using embedding-based reward models. We discussed future avenues for research, aiming to contribute to safer and more effective LLM deployments.
\end{abstract}

\section{Introduction}
Large Language Models (LLMs) have achieved great success on structured tasks like mathematical reasoning and code generation~\citep{guo2025deepseek, jaech2024openai, trinh2024solving} using the technique of Reinforcement Learning (RL). In such a process, rule-based reward functions can be explicitly defined to guide optimization.

On the other hand, in broader applications such as chat bots and general content generators, RL is also an essential technique in aligning LLMs for their safe and successful deployment~\citep{christiano2017deep, ouyang2022training,stiennon2020learning}, and the process is known as Reinforcement Learning from Human Feedback (RLHF). In such a process, reward models serve as a crucial mechanism for quantilizing content values and scaling RLHF~\citep{lambert2024rewardbench, wang2024secrets} --- those models act as proxy evaluators (of human values) during fine-tuning and deployment~\citep{dubey2024llama, dong2024rlhf, wang2024arithmetic}, assessing how well LLM outputs align with human intent. 

\begin{table}[t!]
\fontsize{7.8}{12}\selectfont
\centering
\caption{Training (with $10000$ samples) and evaluation (with $4000$ samples) time comparison for different reward model choices on CPUs and GPUs. (Details of the accelerated workflow are discussed in Section~\ref{sec:motivations}).}
\label{tab:execution_time}
\begin{tabular}{l|l|c|c}
\toprule
\textbf{Reward Models} & \textbf{Example} & \textbf{CPU} & \textbf{Tesla V100} \\
\midrule
Embedding-based & LightGBM & 8s    & -    \\
(Our Position)& 3-layer MLP & 28s   & 33s  \\
\hline
\multirow{2}{*}{LLM-based}       & GPT2 & $>$2h   & 879s \\
                                    & Gemma2B (LoRA R=8) & -     & 4755s \\
\bottomrule
\end{tabular}\vspace{-0.3cm}
\end{table}


Despite significant progress, reward model training remains challenging due to the scarcity and inaccuracy of annotations, inherent complexity, and variability of human preferences~\citep{lambert2024rewardbench,wang2024secrets,gao2023scaling}. Prior research has attempted to mitigate these challenges through improved architectures~\citep{wang2024arithmetic, wang2024arithmetic}, customized loss functions~\citep{winata2024metametrics,liu2024skywork}, uncertainty quantification~\citep{lou2024uncertainty,coste2023reward,zhang2024overcoming}, novel comparisons~\citep{sun2024rethinking,yin2024relative}, dataset debiasing techniques~\citep{park2024offsetbias}, and active or online annotation algorithms~\citep{xiong2023gibbs,muldrew2024active,dong2024rlhf}.

Reward modeling is a rapidly evolving research field, but its progress is significantly hindered by the high computational cost of \textbf{training and evaluating reward models}, which in turn poses challenges for reproducibility across different implementations and fair comparisons among methods.

\textbf{In this paper, we argue that building reward models using embedded input can greatly accelerate research in this field.} Specifically, it enhances reproducibility by reducing hardware and computational resource requirements, cutting the cost of training and evaluation, improving training stability, and minimizing the cost of reproduction, hence accelerating the pace of reward model research. Additionally, it opens new avenues for further exploration such as research using the statistical lenses.

\begin{table*}[ht!]
\fontsize{8}{11.8}\selectfont
\centering
\caption{Comparative of LLM-based and Embedding-based Reward Models.}
\begin{tabular}{p{1.75cm}|c|c|p{8cm}}
\toprule
\textbf{Metric} & \textbf{LLM-based RM} & \textbf{Embedding-based RM} & \textbf{Are Embedding-based Reward Models Better (than LLM-based)?} \\
\midrule
Parameter Count & 3M - 3B & 0.6M & Yes \\ \hline
Performance (Section~\ref{sec:RM_from_embeddings}) & Larger LLMs are stronger & Intermediate & Not Always. Embedding-based methods are remarkably better than small LM-based (GPT2, 700M) reward models, and can be better than larger LM-based (Gemma 2B) reward models when annotation quality and quantity are restricted. In more general setups, further research is needed to enhance those methods. \\ \hline
Training Cost (Section~\ref{sec:cheap_train}) & High (GPU, for hours) & Low (CPU, a few minutes) & Yes, they reduce hardware requirements and lower entry barriers of research \\ \hline
Evaluation Cost (Section~\ref{sec:scalable_evaluation}) & High ($10^2\times$ GPU hours) & Low (CPU, a few minutes) & Yes, fast evaluations enable reliable results and comprehensive comparisons.\\ \hline
Inference Cost (Section~\ref{sec:cheap_inference}) & High (GPUs needed) & Low (GPU-free) & Yes, they enable efficient inference-time optimization and support lightweight deployment across varying infrastructures.\\ \hline
Reproducibility (Section~\ref{sec:reproduce_demo})  & Low  & High & Yes, embedding-based reward models have much less vulnerable hyper-parameters and are easy to reproduce. \\
\bottomrule
\end{tabular}
\label{tab:reward_model_comparison}
\end{table*}

This position paper is structured as follows: In Sec.~\ref{sec:RM_from_embeddings}, we present and compare embedding-based reward models with conventional LLM-based reward models, where general-purpose  LLMs with value heads are optimized to serve as value predictors. In Sec.~\ref{sec:motivations}, we elaborate on the motivations of training reward models using embeddings as inputs, and demonstrating its advantages in practice --- high reproducibility with low cost associated with training (Sec.~\ref{sec:cheap_train}), evaluation (Sec.~\ref{sec:scalable_evaluation}), and inference (Sec.~\ref{sec:cheap_inference}). 
In Sec.~\ref{sec:reproduce_demo}, we demonstrate our positions through an efficient reproduction of existing reward modeling research.
Sec.~\ref{sec:future_works} explores open questions and future research opportunities in this domain. Lastly, Sec.~\ref{sec:alternative_views} provides alternative perspectives to enhance the comprehensiveness of this position paper.

\section{Reward Models with Embedding Inputs}
\label{sec:RM_from_embeddings}

\begin{figure}[h!]
    \includegraphics[width=1.0\linewidth]{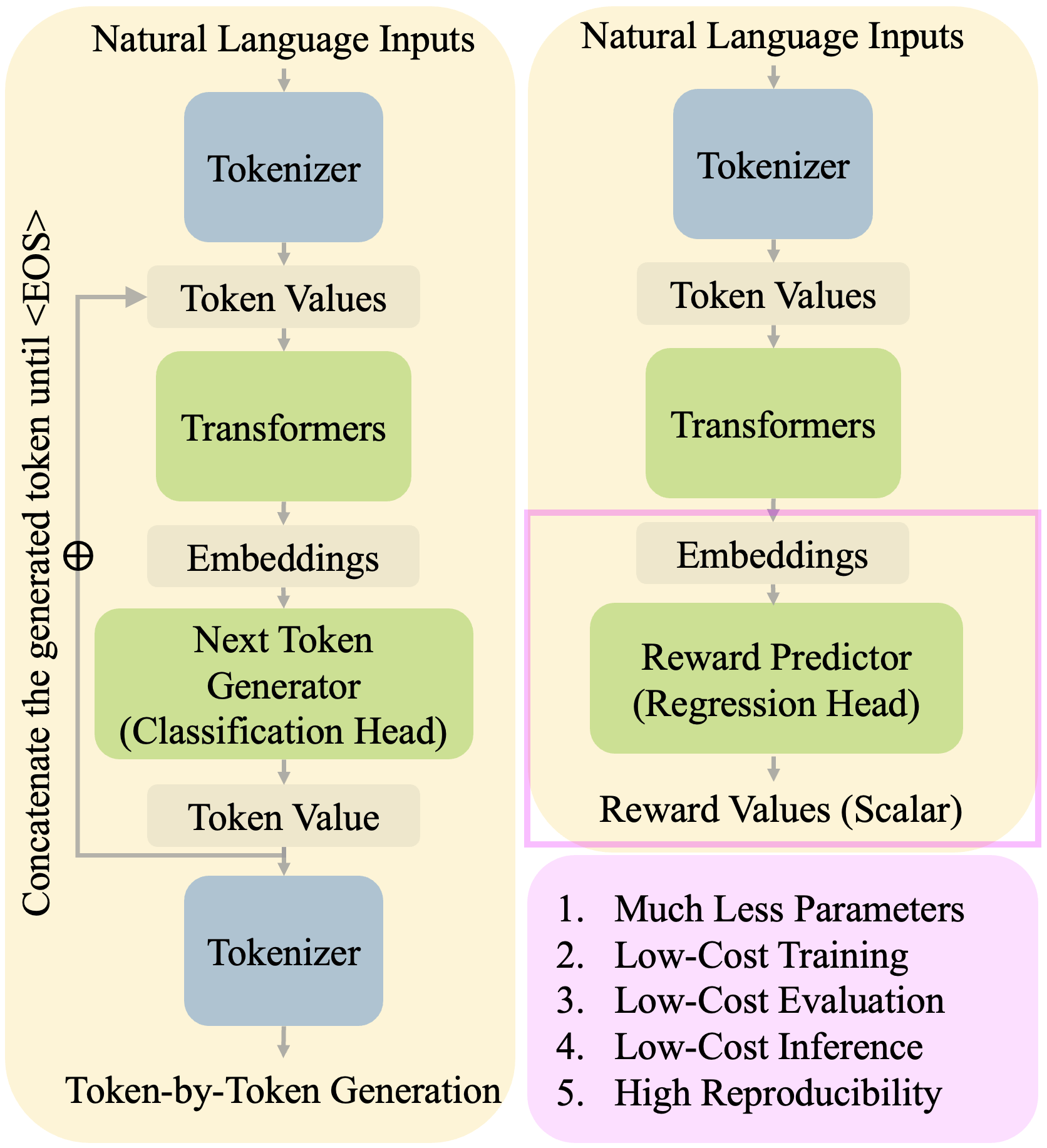}
    \vspace{-0.39cm}
    \caption{In reward model research, using embeddings as input (i.e., focusing on the pink box) brings the following benefits:
    1. there are much less parameters in those reward models;
    2. it has a much lower training cost than using LLM-based reward models;
    3. it has a much lower evaluation cost as compared to the LLM-based reward models; 
    4. it minimizes the inference-time cost by generating embeddings as by-products in language generation;
    5. research using embedding-based reward models are highly reproducible due to the low computational demand, high training stability, and minimal hardware requirement.}
    \label{fig:teaser}
\vspace{-0.3cm}
\end{figure}

\subsection{Alternatives to LLM-based Reward Models}

In this section, we explore the use of embeddings as inputs for reward modeling and contrast this approach with traditional methods employing natural language inputs. Figure~\ref{fig:teaser} illustrates the key differences: green boxes represent trainable parameter groups, while gray boxes denote intermediate variables. The left panel depicts the processing of natural language inputs by LLMs for generation tasks, whereas the right panel shows their use in LLM-based reward models for quality evaluation.

When LLMs equipped with replaced value heads are utilized for reward modeling, only a minimal number of parameters are removed. Consequently, these models retain a substantial degree of parameter freedom, making them large and computationally demanding. For example, training a 2B-parameter model using LoRA~\citep{hu2021lora} on a Tesla-V100 GPU with a typical alignment dataset of 10,000 samples approximately requires two hours. Additionally, the training process involves numerous hyperparameters that can significantly influence the outcomes.

Conversely, given the aim to evaluate natural language content effectively, and recognizing that the embedding space encapsulates a rich representation of the input both before and during the LLM era—as evidenced in tasks ranging from classification to more complex applications~\citep{mikolov2013efficient,pennington2014glove,devlin2018bert,kiros2015skip,cer2018universal,brown2020language} --- employing only embeddings as inputs presents a viable alternative. Recent studies have demonstrated the efficacy of this approach in constructing reward models for prompt evaluation in mathematical reasoning tasks and for assessing the safety and helpfulness of LLM-generated contents~\citep{sun2023query,sun2024rethinking}. Typically, these models require only 1 to 5 minutes of training time on CPU-only machines.

Moreover, as \textbf{embeddings are generated as by-products during the language generation process, utilizing them for reward models imposes no additional computational overhead}. To have a comprehensive understanding of this alternative method, we present experimental results that empirically compare the two approaches' performance in reward modeling.


\subsection{Empirical Comparisons}
\paragraph{Reward Model Sizes} In the extant literature on reward models, LLMs typically range from 3M to 3B parameters, with specific instances such as \citet{coste2023reward} employing models between 14M and 1.3B parameters, \citet{ahmed2024scalable} using a 1.3B model, and \citet{gao2023scaling} exploring models from 3M to 3B parameters. By contrast, embedding-based methods, such as a typical 3-layer MLP with $2048$-dimensional input embeddings and $256$ hidden units, utilize fewer than 0.6M parameters. We also consider lightGBM models in our demonstrative experiments given their wide success and remarkable stability~\citep{ke2017lightgbm,grinsztajn2022tree,sun2023query}.

\paragraph{Data Generation Processes} We use the Anthropic-HH dataset, which includes the \texttt{Helpful} and \texttt{Harmless} alignment tasks to assess the efficacy of various reward model approaches~\citep{bai2022training}. The dataset contains $40000$ prompts for each task. To ensure reproducible and reliable comparisons, we use golden reward models as proxy annotators following established workflows in the literature~\citep{xiong2023gibbs,dong2024rlhf,dong2023raft,gao2023scaling,yang2024rewards}. We consider three LLMs --- Gemma-2B and -7B \citep{team2024gemma}, and LLaMA3-8B \citep{touvron2023llama} --- to generate responses. For each prompt, we generate $10$ responses and randomly select $N$ pairs for preference annotation using the golden reward models. We use Gemma2B to generate embeddings for the embedding-based reward models. This approach ensures that our evaluation accurately reflects the preferences of the golden reward model, thereby minimizing bias.

\paragraph{Annotation Quality and Quantity Control} In preference generation, the quality of annotations is often limited by the capabilities of the annotators \citep{sanderson2010user,stewart2005absolute,guest2016relative,wang2024secrets}. We apply the location-scale function class to describe annotation noise \citep{sun2024rethinking}, positing that closer values yield noisier personal preferences. We examine three annotation quality scenarios: 
\begin{enumerate}[nosep,leftmargin=*] 
\item \textbf{Low annotation quality}: high error rates (approximately $45\%$), offering minimal informative value in the preference annotations; 
\item \textbf{Medium-Low annotation quality}: error rates around $40\%$; 
\item \textbf{Medium-High annotation quality}: error rates around $30\%$; 
\item \textbf{High annotation quality}: error rates are about $5\%$. \end{enumerate}
In addition to quality, we also explore the impact of varying annotation quantities, considering the number of annotated preference pairs ranging from $500$ to $10000$.

\begin{figure*}[t!]
    \centering
    \includegraphics[width=1.0\linewidth]{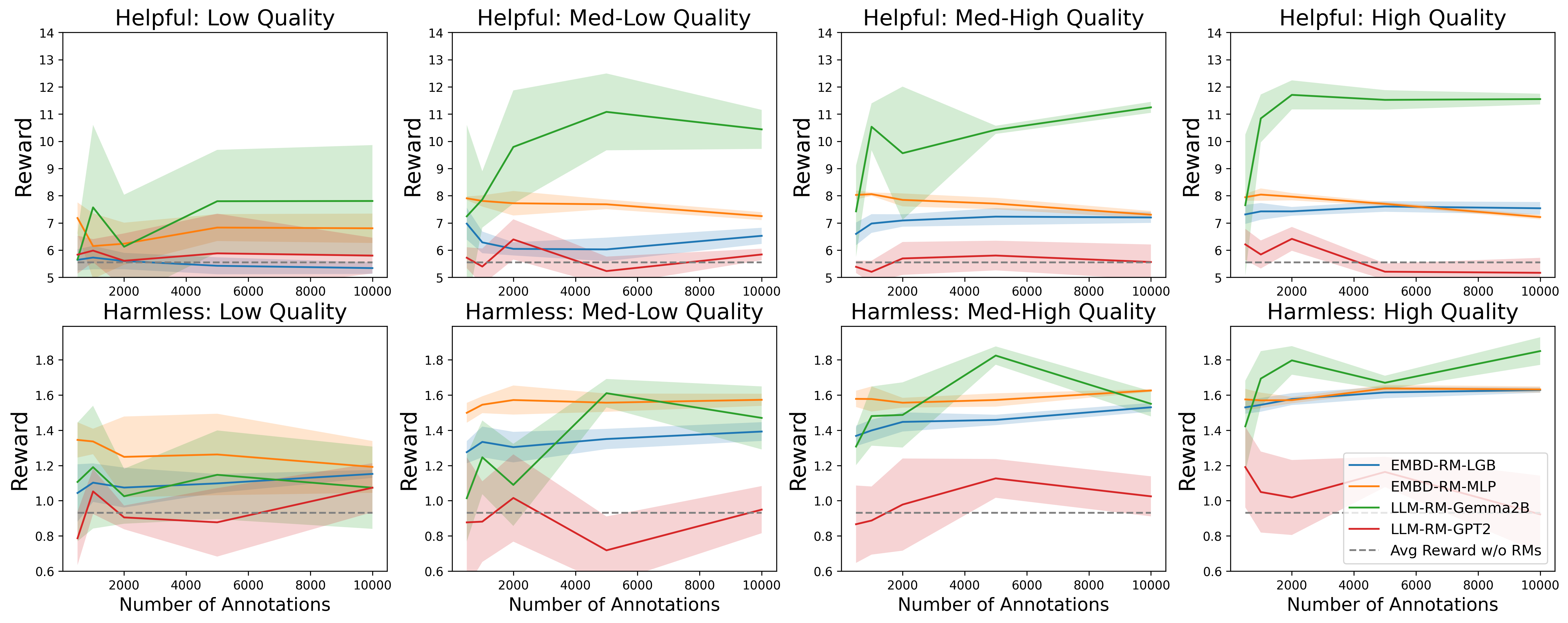}\vspace{-0.35cm}
    \caption{\small  Comparing performances of Embedding-based RM  with LLM-based RMs. The Embedding-based RMs demonstrate high learning stability and strong performance as compared to LLM-based RMs, but are much cheaper to train and evaluate, and more scalable in inference time. Results are from the Gemma 2B model. Additional results using the Gemma 7B and LLaMA3 8B models are presented in Appendix~\ref{appdx:more_results}}
    \label{fig:performance_with_embeddings}\vspace{-0.25cm}
\end{figure*}

Results are presented in Figure~\ref{fig:performance_with_embeddings}. The following observations can be drawn from the analysis:

\begin{itemize}[nosep,leftmargin=*]
\item Generally, embedding-based methods exhibit significantly lower variance and higher stability during training compared to other models. 
\item Embedding-based methods consistently outperform smaller language models (such as LLM-RM-GPT2) across all evaluated scenarios. 
\item In conditions of low annotation quality, embedding-based methods demonstrate performance that is superior to or comparable with LLM-based reward models. 
\item With limited annotation quantities, embedding-based methods also show superior or comparable performance to LLM-based reward models. 
\item On the \texttt{Harmless} dataset, embedding-based reward models consistently match the performance of LLM-based reward models. 
\item On the \texttt{Helpful} dataset, however, embedding-based reward models underperform relative to Gemma2B-based LLM reward models, which benefit more significantly from increases in annotation quality and availability. \end{itemize}

These datasets will be made available as public assets to facilitate future research in reward modeling. Details on the scalable evaluation procedure will be provided in Section~\ref{sec:scalable_evaluation}.

\section{Motivations of Using Embeddings as Reward Model Inputs}
\label{sec:motivations}
\subsection{Reproducibility: Foundation of Research}
\label{sec:cheap_train}
Reproducibility is the foundation of scientific research. In the study of reward modeling, the ability to replicate results across different studies is essential for evaluating theoretical and practical contributions. Nonetheless, the reproduction of LLM-based reward model research often faces considerable obstacles, such as vulnerability to many sensitive hyperparameters, the necessity for large memory GPUs, large training instability, and extensive computational demands associated with slow training processes. These challenges can make the replication of existing works extremely challenging --- if not unfeasible --- for many of the research communities.

\begin{table*}[h!]
\begin{lstlisting}
# Load Training Data
train_embeddings, train_rewards = load_embd_data(task='Harmless', split='train')
print(train_embeddings.shape) 
### (40000, 10, 2048)
print(train_rewards.shape) 
### (40000, 10, 1)

# Load Testing Data
test_embeddings, test_rewards = load_embd_data(task='Harmless', split='test')
print(test_embeddings.shape) 
### (2000, 500, 2048)
print(test_rewards.shape) 
### (2000, 500, 1)

# Generation of Pairwise Comparisons
train_comparisons, train_labels = pair_annotate(train_embeddings, train_rewards)

# Train Embedding-based Reward Model (e.g., use a Bradley-Terry MLP)
reward_model = BT_MLP()
reward_model.fit(train_comparisons, train_labels)

# Make Predictions with the Reward Model on Testset
rm_predictions = reward_model.predict(test_embeddings)
print(rm_predictions.shape) 
### (2000, 500, 1)

# Calculate Evaluation Metrics on Testset
bon_500 = calc_bon(rm_predictions, test_rewards, N=500)
spearmanr = calc_spearmanr(rm_predictions, test_rewards)
\end{lstlisting}
\label{algo}\vspace{-0.4cm}
\end{table*}

As a consequence, in new research, if a method lacks systematic comparisons with established methods due to the above challenges, or its efficacy can not be verified through repeated and statistically significant trials, the results may be unfounded. 

The utilization of embedding-based reward models offers several advantages: \begin{enumerate}[nosep,leftmargin=*] 
\item \textbf{Reward Model Research without GPUs:} Conducting research and reproducing reward model research using embedding-based methods do not necessitate advanced, large-memory GPUs, thereby democratizing access to state-of-the-art research methods and facilitating the validation of novel algorithms by a wider academic community. 
\item \textbf{Lower Computational Requirements for Statistical Significance:} In embedding-based reward model research, the computational overhead is lower not only because of the much cheaper model training process but also for the more consistent results across multiple runs. And there are much less vulnerable hyperparameters that may drastically affect the results. This efficiency enables researchers to rapidly prototype, validate ideas, and innovate based on reliable conclusive empirical observations, maximally isolating the source of gains from complex LLM-based reward modeling systems, thereby accelerating the cycle of scientific discovery and validation in the field. 
\item \textbf{Data Standardization and Scalability:} In embedding-based reward model research, it is possible to create and share a standardized, publicly accessible dataset that includes multiple language models' generations (generality among models), contains a large number of samples (sufficient data for training), flexibly simulate annotation strategies (to stress test methods), and cost-efficient evaluation process.
\end{enumerate}

All of those aspects encourage reproducible research in embedding-based reward modeling, and thereby accelerate the pace of discoveries in the area.

\subsection{Scalable Evaluation with Embedding-based Reward Models}
\label{sec:scalable_evaluation}
In addition to the high computational costs of training, LLM-based reward modeling faces significant challenges in evaluation time and expense. Specifically, reward models are tasked with evaluating test-time generations to differentiate superior responses from inferior ones. Previous research has primarily utilized two metrics for this purpose: LLM-as-a-Judge and evaluation using open-sourced golden reward models~\citep{dong2023raft,dong2024rlhf}.

\textbf{High Cost in LLM-as-a-Judge Evaluation.}
The LLM-as-a-Judge evaluation, which often involves calling commercial APIs, can be prohibitively expensive for even medium-sized datasets. For example, in a study involving $3$ different language models and $2$ datasets, evaluating a proposed method using $2000$ test samples --- each comparison truncated to $1024$ tokens --- through the GPT-3.5 API incurs a cost of $20$ US dollars \textbf{per experiment}, and this cost will be amplified by the number of individual run of the experiment. Compounding such an issue, those results are not reusable.

Moreover, recent findings have exposed potential cheating behavior in LLM-as-a-Judge evaluations~\cite{zheng2024cheating}, further compromising the reliability of this costly method and challenging its feasibility as a community standard.




\textbf{Cost in Golden Reward Model Evaluation.} While the Golden Reward Model Evaluation avoids the use of commercial APIs, making it more accessible and economical for researchers, it still imposes substantial computational demands. For example, evaluating the aforementioned test case necessitates the LLM-based RM to process $12000$ pairs of sequences. In the more computationally intensive best-of-N evaluations, a typical study with $N=500$~\citep[KL divergence approximately 5 Nats,][]{gao2023scaling} requires \textbf{6 million forward passes}. Completing these passes using 2B-parameter LLMs on Tesla V100 GPUs can consume over 100 GPU hours. It is worth noting that this cost is associated with a single experiment setup and a single experimental trial.

\textbf{Cheap and Fast Evaluation with Embedding-based Reward Models.}
In contrast, embedding-based reward models leverage fixed embeddings, allowing for the preparation of a \textbf{standardized test dataset that is reusable across various methods}. For instance, in the scenario described above, we only need to generate the embeddings and golden rewards for the $500$ test responses on each prompt \textbf{once}. These embeddings and rewards are then reusable for any embedding-based reward model evaluation.

We have implemented such a preprocessing step, resulting in a dataset asset that includes $500$ responses for each test prompt. To provide a clearer understanding of the dataset prepared for embedding-based reward modeling, we provide the \grayboxtext{pseudo-code in the box} in page \pageref{algo} for illustration.

In such a use case, the computationally intensive step of embedding generation is completed during data preparation. Subsequently, the evaluation involves merely processing test tensors of shape $(2000, 500, 2048)$ through the reward model --- a task that typically concludes within a minute. This efficiency highlights the practicality of our embedding-based reward modeling framework, which significantly simplifies and accelerates the evaluation of reward models and improves its reliability.





\subsection{Scalable Inference-Time Optimization}
\label{sec:cheap_inference}
\begin{figure}[h!]
\vspace{-0.15cm}
    \includegraphics[width=1.0\linewidth]{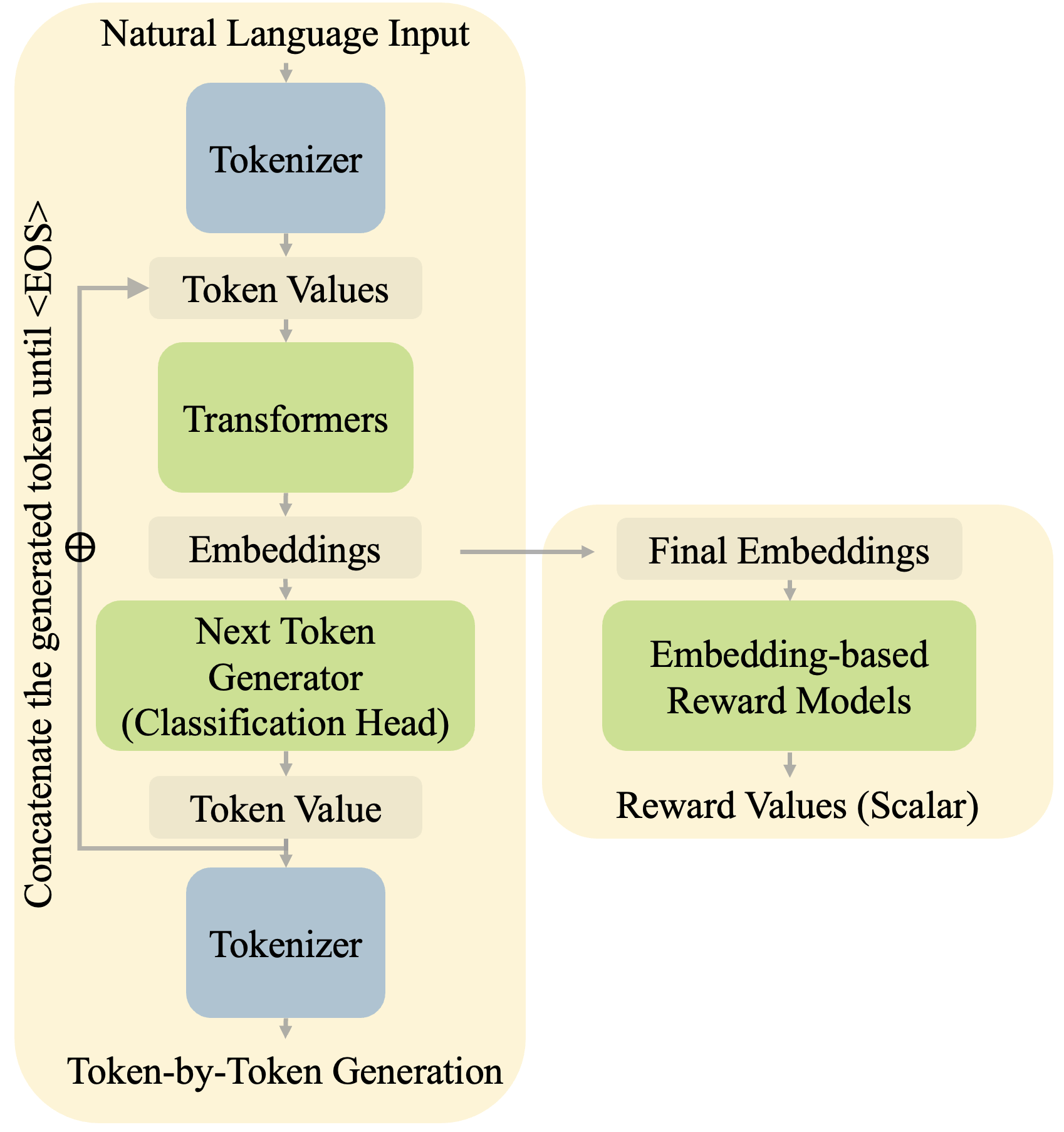}
    \vspace{-0.6cm}
    \caption{\small \textit{The inputs of embedding-based reward models are by-products of language model generation.} Unlike conventional LLM-based reward models that require another LLM forward pass for inference time evaluations, embedding-based models alleviate the memory challenge and facilitate inference time optimization for LLM-free service providers. These providers, who rely on third-party LLM services via APIs rather than hosting large models locally, can efficiently perform inference time optimization using only embeddings.}  \vspace{-0.1cm}
    \label{fig:fast_inference}
\end{figure}

In this section, we elucidate an additional advantage of embedding-based reward models --- enhancing the inference-time optimization efficiency. With embedding-based reward models, language generation and evaluation require only a single LLM forward pass. Although this may appear to reduce computation time by less than half, it significantly lowers the computational burden in evaluation by shifting from hosting an LLM (reward model) to a much simpler and smaller model. This is particularly beneficial for API-based service providers who previously could not perform inference-time optimization due to the high computational demands of running LLM-based reward models locally. With embedding-based reward models, they are now able to efficiently evaluate the quality of generated content and potentially enhance user experience through inference-time optimization (e.g., prompting optimization and re-generation). The workflow of using embedding-based reward models in inference is visualized in Figure~\ref{fig:fast_inference}.


\section{Case Study: Efficient Reproduction of Reward Model Ensemble Papers}
\label{sec:reproduce_demo}




In this section, we replicate the findings from prior research on mitigating overoptimization in reward models through ensemble methods, as discussed in \citep{coste2023reward, ahmed2024scalable}, using our proposed embedding-based reward modeling framework.

To validate the principal finding in those works that ensembles can alleviate reward model overoptimization, we train $10$ LightGBM models \citep{ke2017lightgbm} using default hyperparameter settings, alongside an MLP-based implementation with $256$ hidden units. We assess the performance of these ensemble reward models by averaging predictions across the $10$ models. Experiments are repeated with $5$ independent runs to draw statistically significant conclusions.

Our experiments are conducted on a machine equipped with a 128-core \texttt{Intel(R) Xeon(R) Platinum 8336C CPU @ 2.30GHz}. Our experimental setup encompasses $2$ different models (MLP and LightGBM), $2$ tasks, and build reward models for $3$ LLMs (Gemma 2B, Gemma 7B, LLaMA3 8B). We explore $4$ different annotation quality scenarios and $5$ levels of annotation quantity, ranging from $500$ to $10000$.
\begin{figure}[h!]
    \includegraphics[width=1.0\linewidth]{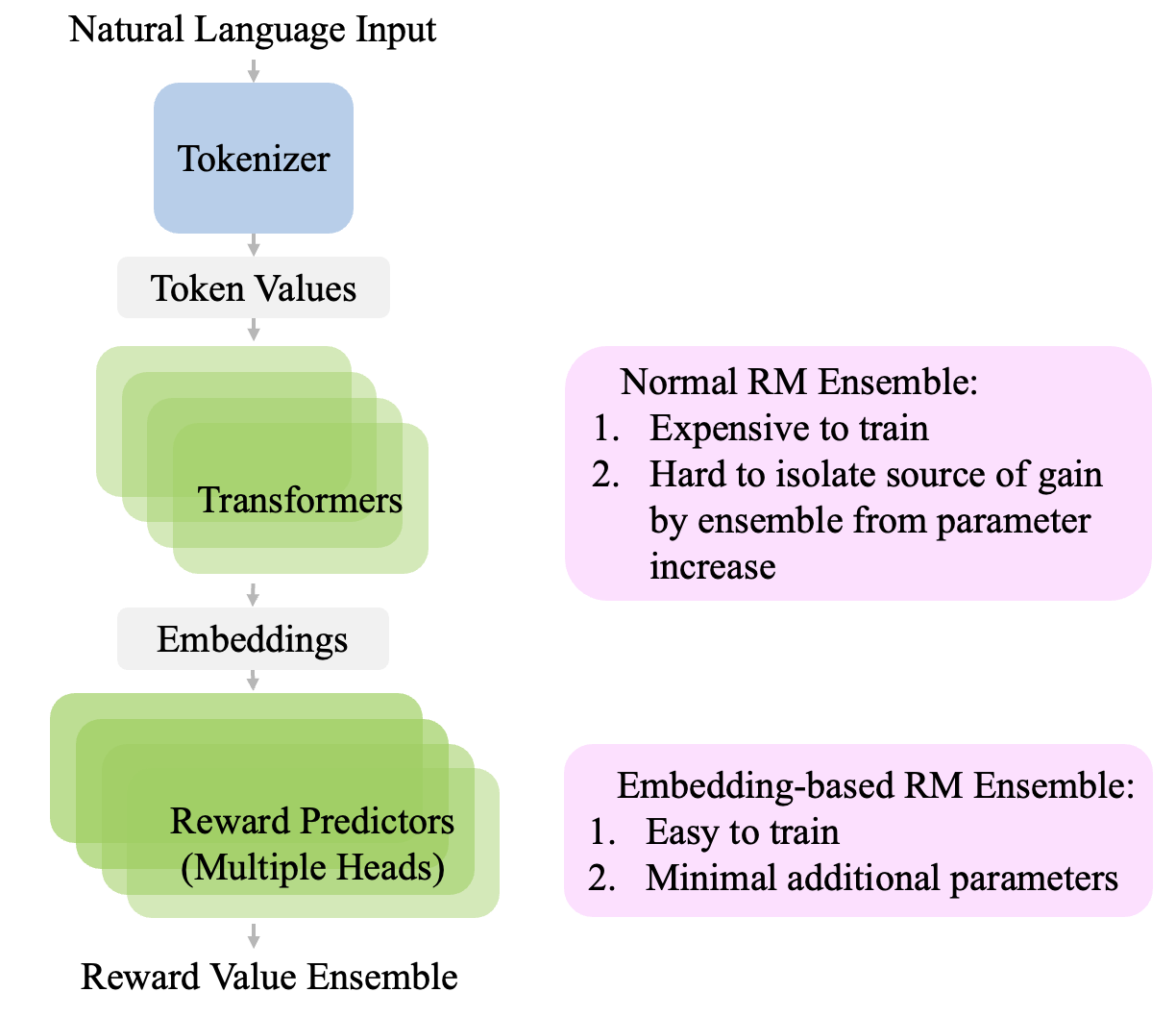}
    \vspace{-0.6cm}
    \caption{\small Using embeddings as inputs in a lightweight reward model ensemble practice to mitigate reward overoptimization. Reproduction of prior findings across over $12000$ configurations can be completed in less than 1 day using CPU-only resources.}\vspace{-0.35cm}
    \label{fig:rm_ensemble_illu}
\end{figure}
\begin{figure*}[t!]
    \centering
    \includegraphics[width=1.0\linewidth]{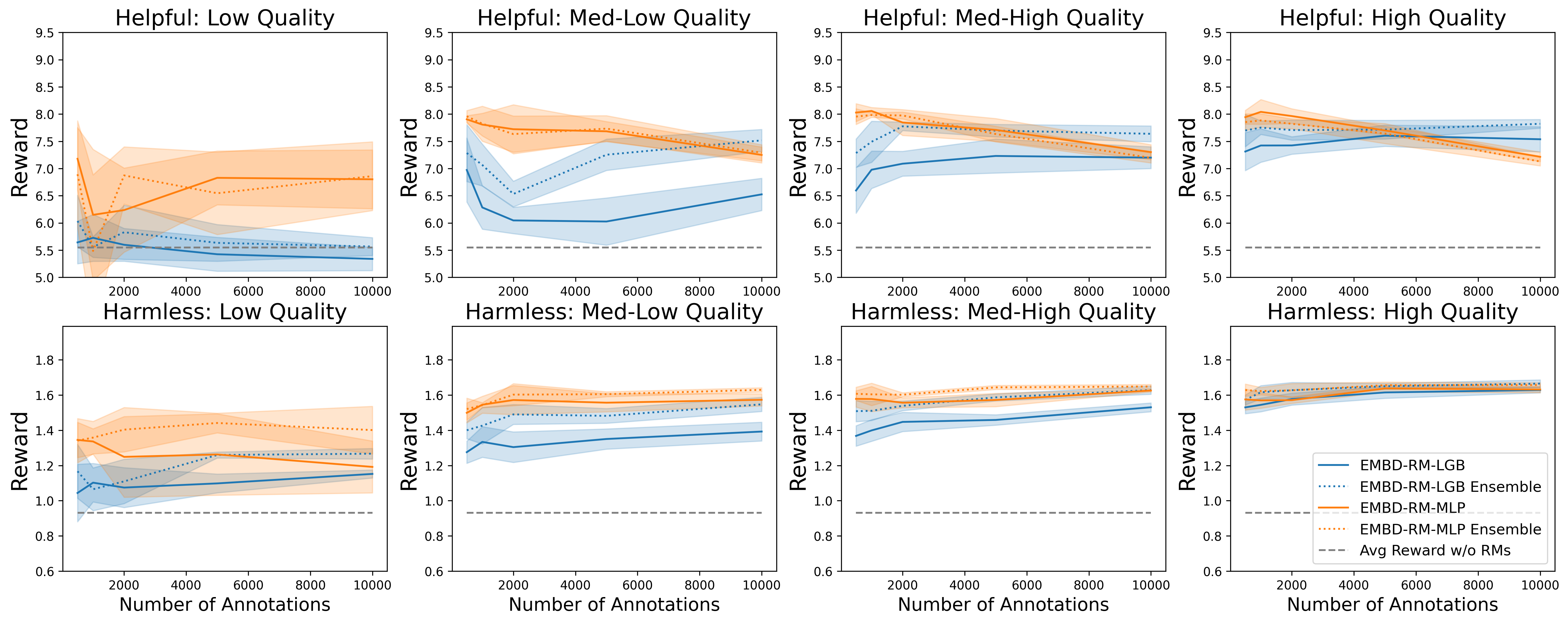}\vspace{-0.65cm}
    \caption{\small Reproduction of reward model ensemble papers using embedding-based reward models. Additional results using the Gemma 7B and LLaMA3 8B models are presented in Appendix~\ref{appdx:more_results}}
    \label{fig:reward_model_ensemble_gemma2b}\vspace{-0.35cm}
\end{figure*}

In total, we train and evaluate $12000$ models. Using the CPU server, training and evaluating the $6000$ LightGBM models takes 4.9 hours, while the $6000$ MLP models require 17.3 hours. In total, these $12000$ experimental configurations are completed within $1$ single CPU day.

Finally, unlike prior research, our investigation into reward model ensembles using embeddings as inputs clarifies that the observed enhancements stem from conservative modeling approaches, rather than from the scaling laws typical of LLM-based reward models \citep{gao2023scaling}. These distinctions are visually demonstrated in the case study illustrated in Figure~\ref{fig:rm_ensemble_illu}. 

Figure~\ref{fig:reward_model_ensemble_gemma2b} shows the results from our efficient reproduction. We observe significant performance improvements when using ensemble methods in reward modeling, thereby verifying the principal findings of \citet{coste2023reward} and \citet{ahmed2024scalable} within embedding-based reward model setups. Notably, the efficacy of the reward model ensemble diminishes as annotation quality improves (i.e., when the error rate is less than $5\%$), and we observe the LightGBM reward models generally get larger performance gains from the ensemble.

\section{Call for Contributions}
\label{sec:future_works}
\subsection{Contributing to Public Embedding Assets}
In this position paper, we have demonstrated the advantages of embedding-based reward models. We successfully reproduced the findings of a reward model ensemble study with $12,000$ experiment runs in just one day using only CPU resources, highlighting the efficiency of our approach. However, it's important to note that this workflow is feasible only when embeddings from LLM generations are available for both training and testing datasets.

In conventional LLM-based reward model research, LLM generations have not been regarded as critical public assets in reward model research, primarily because evaluating these generated contents requires nearly as much computational effort and hardware resources as producing them.

In contrast, our embedding-based reward model framework enables researchers with access only to CPU resources to participate in this field. This inclusivity relies on the availability of embedding assets, contributed by researchers with access to more powerful GPU resources.

For our studies, we utilized the \texttt{Anthropic-HH} dataset and $3$ different LLMs, enabling us to release all corresponding embeddings and their evaluations as public assets for future research. However, given the rapid advancements in general-purpose LLMs, this alone is not enough. \textbf{We encourage more contributions from the community to enrich these assets.}

Moreover, an added benefit of this approach is its environmental impact. By making these assets reusable, other researchers do not need to expend computational and electrical resources to regenerate training and testing samples. This not only accelerates research but also significantly reduces the environmental burden associated with the extensive use of computational resources in large-scale model training and evaluation.





\subsection{Representation Learning: Searching for General Purpose Reward Embeddings}
Current language model embeddings are primarily designed and optimized for text generation. \textbf{While they can be repurposed as inputs for reward models, as demonstrated in this paper, there remains significant room for improvement.} Our experiments indicate that fine-tuning LLM-based reward models, though computationally expensive, can yield superior performance when provided with rich and clear annotation signals.

Given the advantages of embedding-based reward models outlined in this paper, developing better general-purpose reward embeddings represents a promising orthogonal direction for advancing reward model research. 

To link with another important research avenue of the generative reward modeling, where the token generation capabilities of LLMs are directly leveraged for value prediction~\citep{mahan2024generative,zhang2024generative} or used as a regularization mechanism in LLM-based reward model learning\citep{yang2024regularizing}. Their key insight is that \textit{generation ability can enhance performance in discriminative tasks}. In contrast, the question of how to leverage reward modeling information to learn general-purpose discriminative embeddings remains relatively underexplored. Notable exceptions include efforts to merge multiple preference datasets~\citep{dong2024rlhf}. However, \citet{sun2024off} found that combining offline generations with online annotations can be harmful to reward model training. Another related challenge in reward modeling is known to be the alignment tax~\citep{lin2024mitigating}, and how to balance multiple objectives~\citep{yang2024rewards,zhou2023beyond}, and ideal general reward embedding should be able to capture multiple aspects of the responses.




\subsection{Flash Back of Classic Statistics}
Back in the early days of statistical natural language processing, circa the 1990s to early 2000s \citep[for even earlier history, we refer to][]{jones1994natural}, researchers had quite limited options for features even for simple classification tasks. Simple models (e.g., classification trees) were often accompanied by handcrafted, ad hoc features like bags of words, n-grams, and tf-idf \citep{chowdhury2010introduction}, which are seen as insufficient today. With neural networks, representation learning and model development occurred simultaneously; one can even argue that the success of deep models lies in the success of representation learning \citep{bengio2013representation}. Lightweight statistical learning methods possess good properties that are still relevant today. For instance, it is much less resource-intensive and more stable to fit boosted trees than DNNs. The theoretical properties of generalized linear models, some nonparametric regression, as well as tree models, are well understood for classification, preference learning, and for new tasks like experimental design and active learning. 

In future works, can we get the best of both worlds by combining powerful embeddings from an LLM, together with a solid understanding of classic methods to better advance reward modeling with a gray box approach? Can we develop theories building upon the knowledge of classic methods? --- for instance, under the linear assumption with embeddings, what theoretical properties can we establish, and how can we conduct active learning? There are vast research opportunities lying at the interface between statistics and embedding-based reward modeling.

\section{Alternative Views}
\label{sec:alternative_views}
\paragraph{Success of End to End Training.}

The remarkable success of deep learning is largely attributed to the end-to-end learning capability of deep neural networks~\citep{lecun2015deep,goodfellow2016deep}, which has proven effective across diverse domains, including image processing~\citep{krizhevsky2012imagenet,he2016deep}, natural language processing~\citep{vaswani2017attention,devlin2018bert}, tabular data analysis~\citep{arik2021tabnet}, and time series data~\citep{van2016wavenet,ismail2019deep,ding2020hierarchical}.
Representation learning~\citep{bengio2013representation} and pre-training methods~\citep{radford2018improving} are typically followed by post-training or fine-tuning procedures to adapt to downstream tasks or datasets~\citep{howard2018universal,raffel2020exploring,radford2021learning}. In the era of large language models, general-purpose pre-trained models have been extensively fine-tuned for a wide range of downstream applications~\citep{brown2020language}, including evaluation tasks such as reward modeling~\citep{perez2022red,ouyang2022training,chang2024survey,lin2023llm}.

\paragraph{Computational Costs are Decreasing Over Time}
As computational costs continue to decrease, future research on reward models may efficiently leverage LLMs or even more powerful foundation models. This could eliminate the need for embedding-based reward modeling approaches, further supporting the case for end-to-end learning.

From this perspective, one could reasonably argue that reward model learning should ultimately adopt an end-to-end approach. The positions proposed in this paper may only remain valid within a limited timeframe. Future advancements in methodology and hardware technology may render them obsolete.

\bibliography{main}
\bibliographystyle{icml2025}

\newpage
\appendix
\onecolumn
\section{More Results}
\label{appdx:more_results}
\paragraph{Performance Comparison: Embedding-based reward models v.s. LLM-based reward models.}
In our main text, we presented the results with the Gemma 2B model when comparing the performance of different reward modeling approaches. We now provide the results using the Gemma 7B and LLaMA3 8B models as complementary empirical supports. The observations concluded in our main test still hold true on those experiment setups.
\begin{figure*}[h!]
    \centering
    \includegraphics[width=0.98\linewidth]{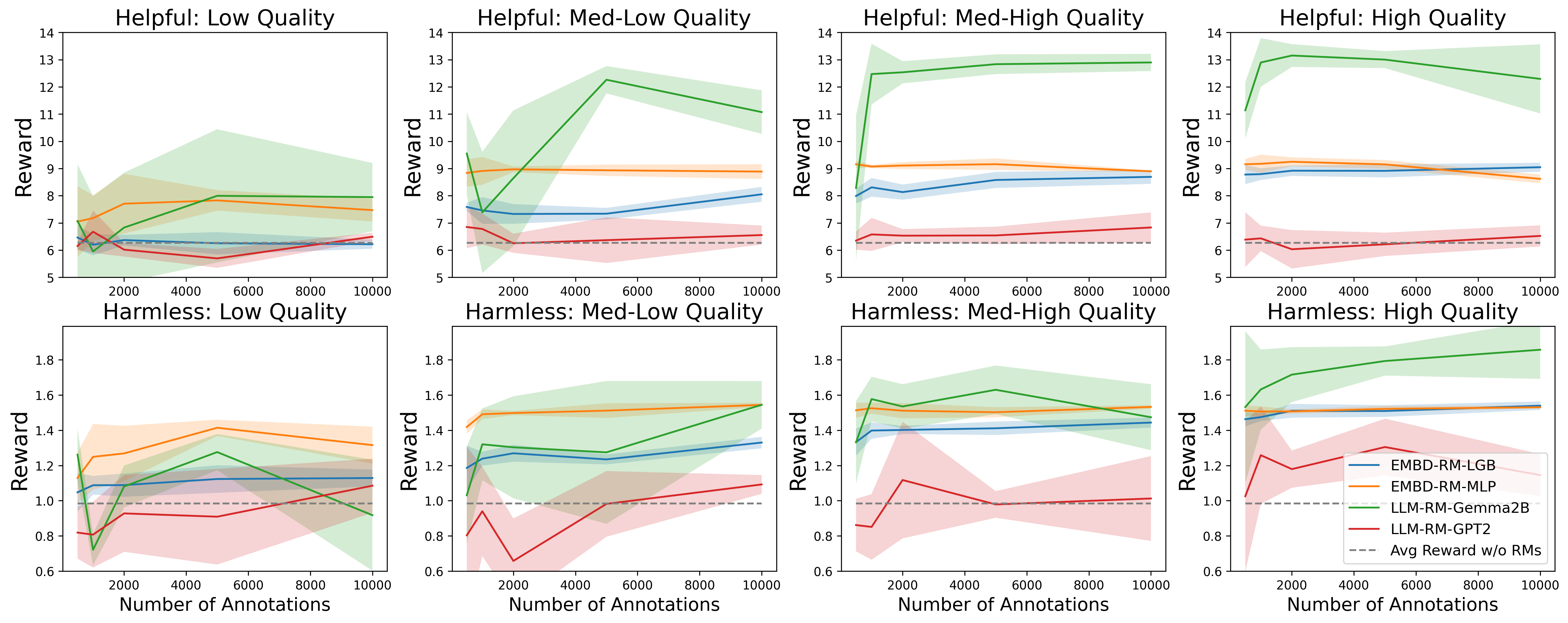}
    \caption{\small  Comparing performances of Embeddings-based RM  with LLM-based RMs using Gemma 7B.}
    \label{fig:performance_with_embeddings_gemma7b}
\end{figure*}

\begin{figure*}[h!]
    \centering
    \includegraphics[width=0.98\linewidth]{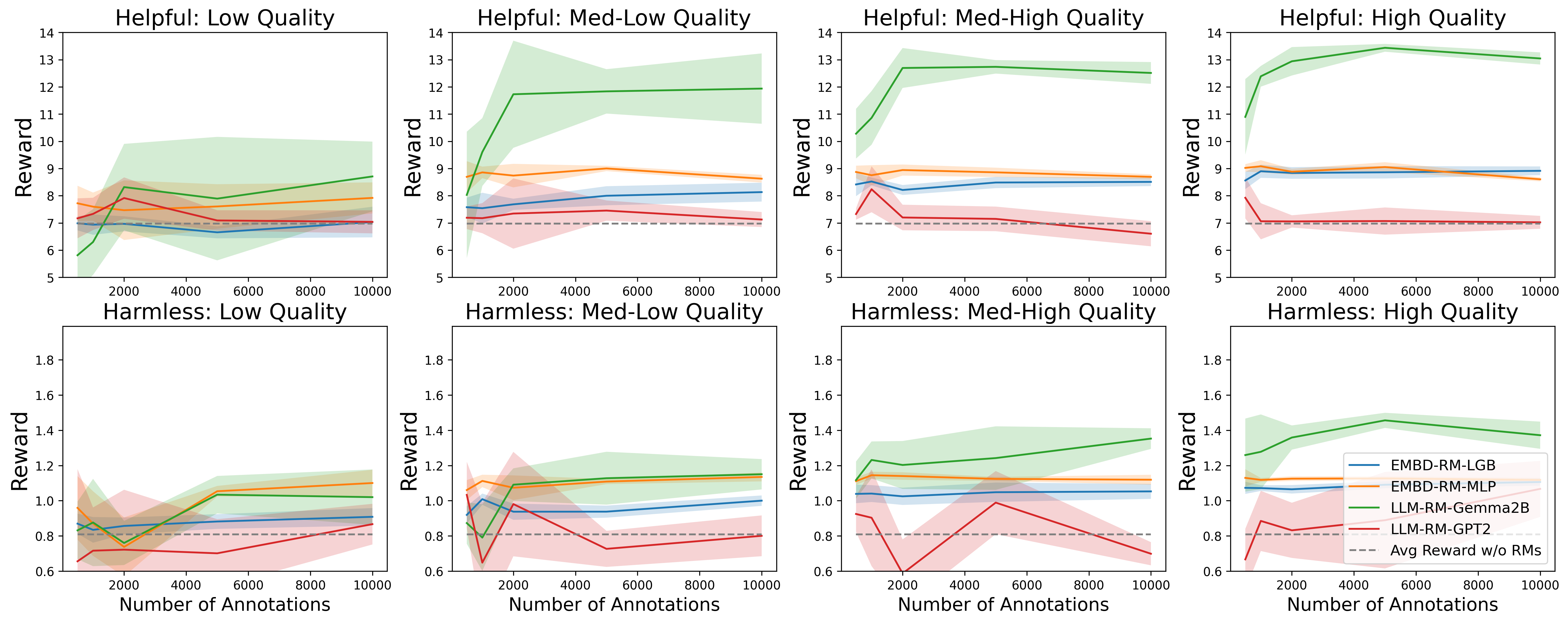}
    \caption{\small  Comparing performances of Embeddings-based RM  with LLM-based RMs using LLaMA3 8B.}
    \label{fig:performance_with_embeddings_llama38b}
\end{figure*}

\newpage
\paragraph{Additional results reproducing reward model ensemble with embedding-based reward models.}
In our main text, we presented the results with the Gemma 2B model when reproducing reward model ensemble papers. We now provide complementary results using the Gemma 7B and LLaMA3 8B models in response generations. The observations concluded in our main test still hold true on those experiment setups.

\begin{figure*}[h!]
    \centering
    \includegraphics[width=1.0\linewidth]{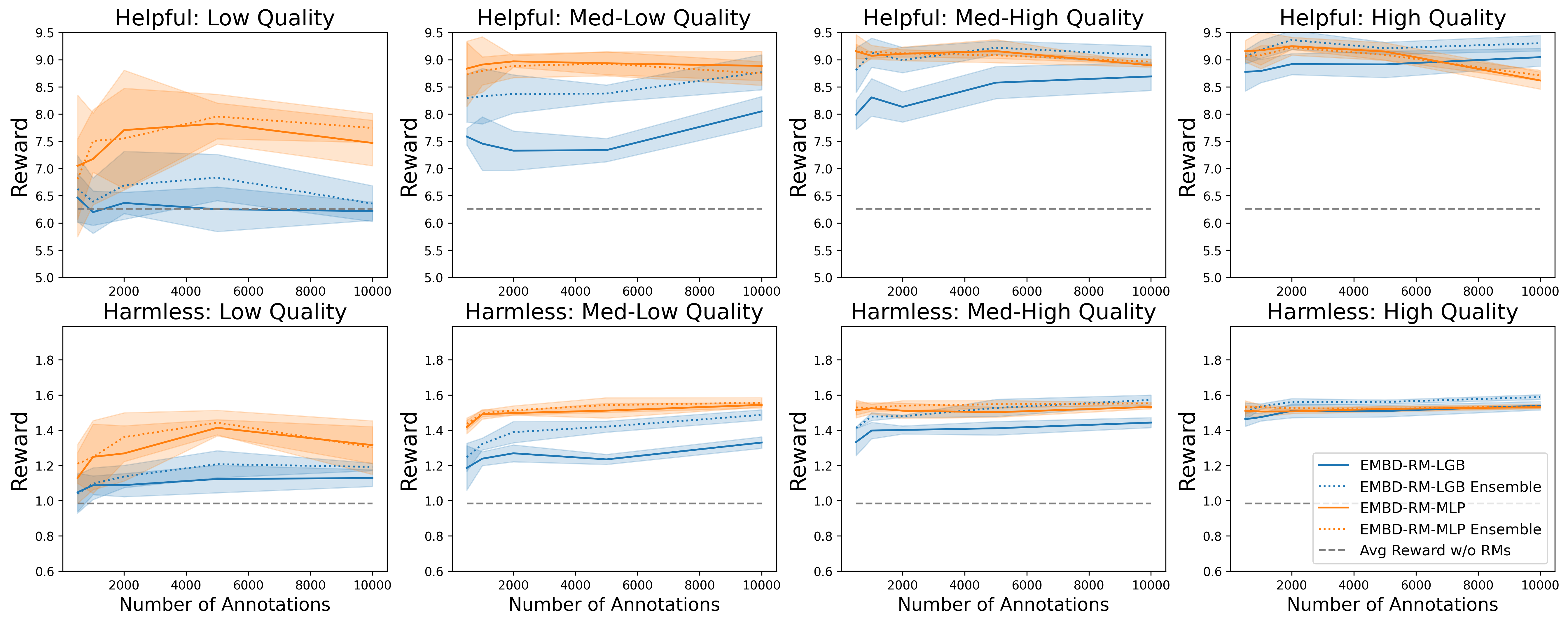}\vspace{-0.35cm}
    \caption{\small Reproduction of reward model ensemble papers using embedding-based reward models. Results on building reward models for Gemma 7B.}
    \label{fig:reward_model_ensemble_gemma7b}\vspace{-0.25cm}
\end{figure*}

\begin{figure*}[h!]
    \centering
    \includegraphics[width=1.0\linewidth]{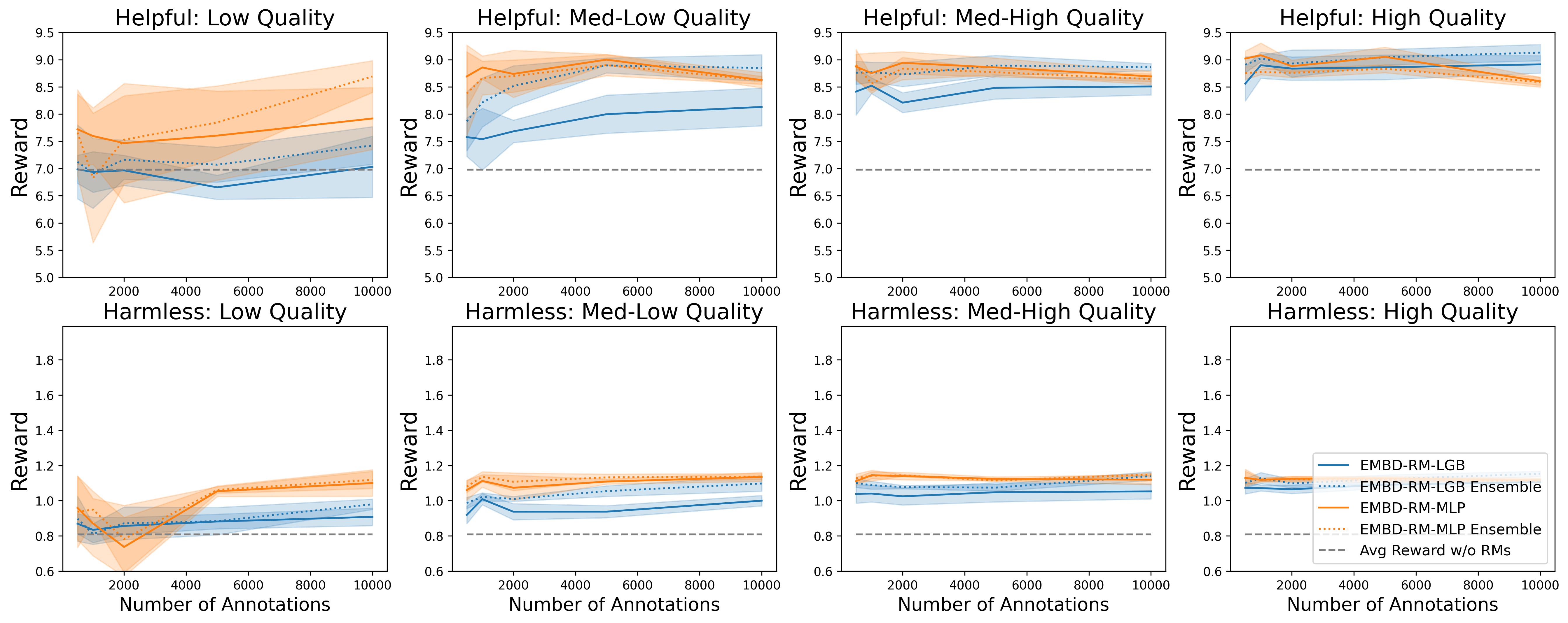}\vspace{-0.35cm}
    \caption{\small Reproduction of reward model ensemble papers using embedding-based reward models. Results on building reward models for LLaMA3 8B.}
    \label{fig:reward_model_ensemble_llama38b}\vspace{-0.25cm}
\end{figure*}

\end{document}